\documentclass[letterpaper]{article} 
\usepackage[]{aaai23}  
\usepackage{times}  
\usepackage{helvet}  
\usepackage{courier}  
\usepackage[hyphens]{url}  
\usepackage{graphicx} 
\usepackage{natbib}  
\usepackage{caption} 
\usepackage{algorithm}
\usepackage{algorithmic}

\usepackage{amsmath}
\usepackage{amsfonts}
\usepackage{subcaption}
\usepackage[dvipsnames]{xcolor}
\usepackage{tikz}
\usetikzlibrary{positioning, arrows, calc, shapes.geometric}
\usepackage{pgfplots}
\usepackage{multirow}
\usepackage{makecell}
\usepackage{soul}

\usepackage{newfloat}
\usepackage{listings}
\DeclareCaptionStyle{ruled}{labelfont=normalfont,labelsep=colon,strut=off} 
\floatstyle{ruled}
\newfloat{listing}{tb}{lst}{}
\floatname{listing}{Listing}
\title{Persistence-Based Discretization for Learning Discrete Event Systems from Time~Series}
\author{
    Lénaïg Cornanguer\textsuperscript{\rm 1}, Christine Largouët\textsuperscript{\rm 2}, Laurence Rozé\textsuperscript{\rm 3}, Alexandre Termier\textsuperscript{\rm 4}
}
\affiliations{
    \textsuperscript{\rm 1}Inria, Univ Rennes, CNRS, IRISA\\
    \textsuperscript{\rm 2}Institut Agro, Univ Rennes, Inria, CNRS, IRISA\\
    \textsuperscript{\rm 3}Univ Rennes, INSA Rennes, CNRS, Inria, IRISA\\
    \textsuperscript{\rm 4}Univ Rennes, Inria, CNRS, IRISA\\
    lenaig.cornanguer@irisa.fr
}

\usepackage{bibentry}

\begin{document}
\nocopyright 

\maketitle

\begin{abstract}
To get a good understanding of a dynamical system, it is convenient to have an interpretable and versatile model of it. Timed discrete event systems are a kind of model that respond to these requirements.
However, such models can be inferred from timestamped event sequences but not directly from numerical data. To solve this problem, a discretization step must be done to identify events or symbols in the time series.
Persist is a discretization method that intends to create persisting symbols by using a score called persistence score. This allows to mitigate the risk of undesirable symbol changes that would lead to a too complex model.
After the study of the persistence score, we point out that it tends to favor excessive cases making it miss interesting persisting symbols.
To correct this behavior, we replace the metric used in the persistence score, the Kullback-Leibler divergence, with the Wasserstein distance.
Experiments show that the improved persistence score enhances Persist's ability to capture the information of the original time series and that it makes it better suited for discrete event systems learning.
\end{abstract}

\section{Introduction}
With the ever-growing type and variety of available sensors, more and more systems can be monitored in real-time.
This monitoring results in the collection of multiple {\em time series}, i.e. sequences of numerical values captured by the sensors.
For example, in a smart city, streets can be monitored through pedestrian counting as well as pluviometers, to study the effect of weather conditions and the time of the day on the number of people walking.

The classical use of the sensor time series is to feed them to machine learning models for tasks such as classification or anomaly detection. 
However, one may also be interested to understand better the dynamical system being monitored by the sensors.
For such goal, an interpretable model is required. 
A good solution is to produce a model in the form of a timed Discrete Event System from the sensor data.
A first difficulty is that there are globally very few approaches learning timed Discrete Event Systems from data. 
The learning of one of these Discrete Event Systems formalisms, Timed Automata (TA), has been well studied with algorithms like RTI+ and TAG \cite{verwer_likelihood_2010, cornanguer_tag_2022}.
However these algorithms take as input sequences of timestamped events, and not time series. 

In order to use these approaches with sensor data, a solution is to discretize the time series before using the automata learning algorithm. 
There is a large literature on time series discretization. 
However, the proposed discretization methods are not designed for automata learning: they may exhibit too frequent consecutive changes of symbols, which would lead to needlessly large and complex automata.
With the idea of minimizing the number of symbol changes, Mörchen at Ultsch \cite{morchen_optimizing_2005} proposed the Persist approach. 
Persist is based on the assumption that the time series reflect the dynamics of an underlying system composed of recurring persisting states and aims at recovering these states in the form of a sequence of symbols issued from the time series discretization.

While the idea of Persist is great, using it in practice for automata learning reveals that Persist may often ``go too far'' by focusing on extreme values leading to a discretized sequence with hardly any symbol changes.
In this paper, through a thorough analysis of the decision criterion of Persist, the persistence score, we identify the source of this behavior leading to a more balanced distribution of the symbols.
Our experiments on numerous real and synthetic datasets demonstrate that the improved persistence score allows to better capture the information of the original time series, and can help in producing interpretable timed automata. 

\section{Motivation and State-of-the-Art}
We consider the problem of converting a numerical time series, a sequence of values measured at regular time intervals~\cite{lin_symbolic_2003}, into a symbolic representation, a sequence of symbols from a finite alphabet. 
Given a time series, $x=\{x_i | x_i \in \mathbb{R}, i = 1, ..., n\}$, the purpose is to provide its discretized version $y=\{y_i | y_i \in \Sigma, i = 1, ..., n\}$ where each symbol $y_i$ is an element of a finite alphabet $\Sigma$. 
Symbolic representation is an efficient way to deal with the inherent dimensionality of time series so that they can be used in a low-dimensional space with data-mining and machine-learning algorithms. 
To address this problem, many approaches have been proposed.
The simplest discretization methods are equal-width (EW) and equal-frequency (EF) interval binning which subdivide continuous ranges into intervals through user specification of width or frequency.
SAX (Symbolic Aggregate approXimation) \cite{lin_symbolic_2003} is another simple and widely implemented method based on the piecewise aggregate approximation (PAA) technique. The time series is divided into equal-size time intervals for which only the mean value is kept. Given the assumption that time series follow a normal distribution, the Gaussian curve is then divided through breakpoints producing equiprobable symbols. SAX requires two parameters, the number of time intervals and the number of symbols.
SAX suffers from limitations such as the normal distribution assumption and user parameters which impact the quality of the results.
Several variants have been proposed to attempt to overcome these limitations. 
Some other approaches like ABBA and fABBA \cite{fABBA_2022} have been developed to capture the shape and the trend of the time series. 
These methods are motivated by different purposes and appear to be best suited for applications dedicated to the analysis of time series such as trend prediction or anomaly detection.
Yet, no discretization method has been specifically proposed with the end goal of analyzing or learning behaviors of dynamical systems.
However, Mörchen and Ultsch~\cite{morchen_optimizing_2005} have proposed a discretization algorithm, Persist, based on an interesting property over time series, called {\em persistence}.
Persist attempts to produce discretized time series with persisting symbols, which would be an advantage for the learning of a timed discrete event system.
Our motivation in this paper is to improve Persist to obtain a more accurate symbolic representation suited for the description of dynamical systems. 

\section{Persist}
Persist \cite{morchen_optimizing_2005} is a method for unsupervised discretization of univariate time series proposed by Mörchen and Ultsch.
It was employed as preprocessing step to find patterns in time series in a language called Time Series Knowledge Representation (TSKR) \cite{morchen_efficient_2007}. 
Persist is based on the assumption that the time series are the reflection of an underlying process that consists of recurring persisting states and it aims to restore these states in the form of symbols in a discretized version of the time series.
Mörchen and Ultsch state that ``if there is no temporal structure in the time series, the symbols [in its discretized version] can be interpreted as independent observations of a random variable according to the marginal distribution of symbols".
Thus, the idea is to look for states showing a persisting behavior by creating symbols whose probability of repetition will be much higher than their probability of appearance.  

To create these symbols, Persist produces a set of breakpoints creating intervals in the value space. Each interval is associated with a symbol $s$ that will replace the numerical values falling within it in the discretized time series. 

The breakpoints are iteratively chosen in a set of candidate breakpoints ($candidates$ in Algorithm \ref{alg:breakpoints}) according to a score called persistence score (described in the next section). The set of candidates is initialized by an equal frequency binning (with a number of bins fixed to 100 by default).
At each iteration, the function \texttt{best\_bp} individually tests every candidate breakpoint added to the already selected breakpoints ($bps$). The candidate increasing persistence score the most is returned with its score. 
Persist stops when no more candidate breakpoint increases the persistence score, finding thereby automatically an adequate number of symbols.

\begin{algorithm}
    \caption{Persist}
    \label{alg:breakpoints}
    \begin{algorithmic}
        \REQUIRE univariate time series $ts$
        \ENSURE a set of breakpoints $bps$
        \STATE $bps = \emptyset$
        \STATE $candidates = \text{equal\_frequency\_binning}(ts, 100)$
        \STATE $score = 0$               
        \STATE $new\_score = 0$
        \STATE $(new\_bp,new\_score) =\text{best\_bp}(ts, bps, candidates)$
        \WHILE{$new\_score < score$}
            \STATE $score = new\_score$
            \STATE $bps = bps \cup new\_bp$
            \STATE $candidates = candidates - {new\_bp}$
            \STATE $(new\_bp,new\_score) =\text{best\_bp}(ts, bps, candidates)$
        \ENDWHILE
        \RETURN $bps$
    \end{algorithmic}
\end{algorithm}

\begin{figure}[t]
\includegraphics[width=.95\linewidth]{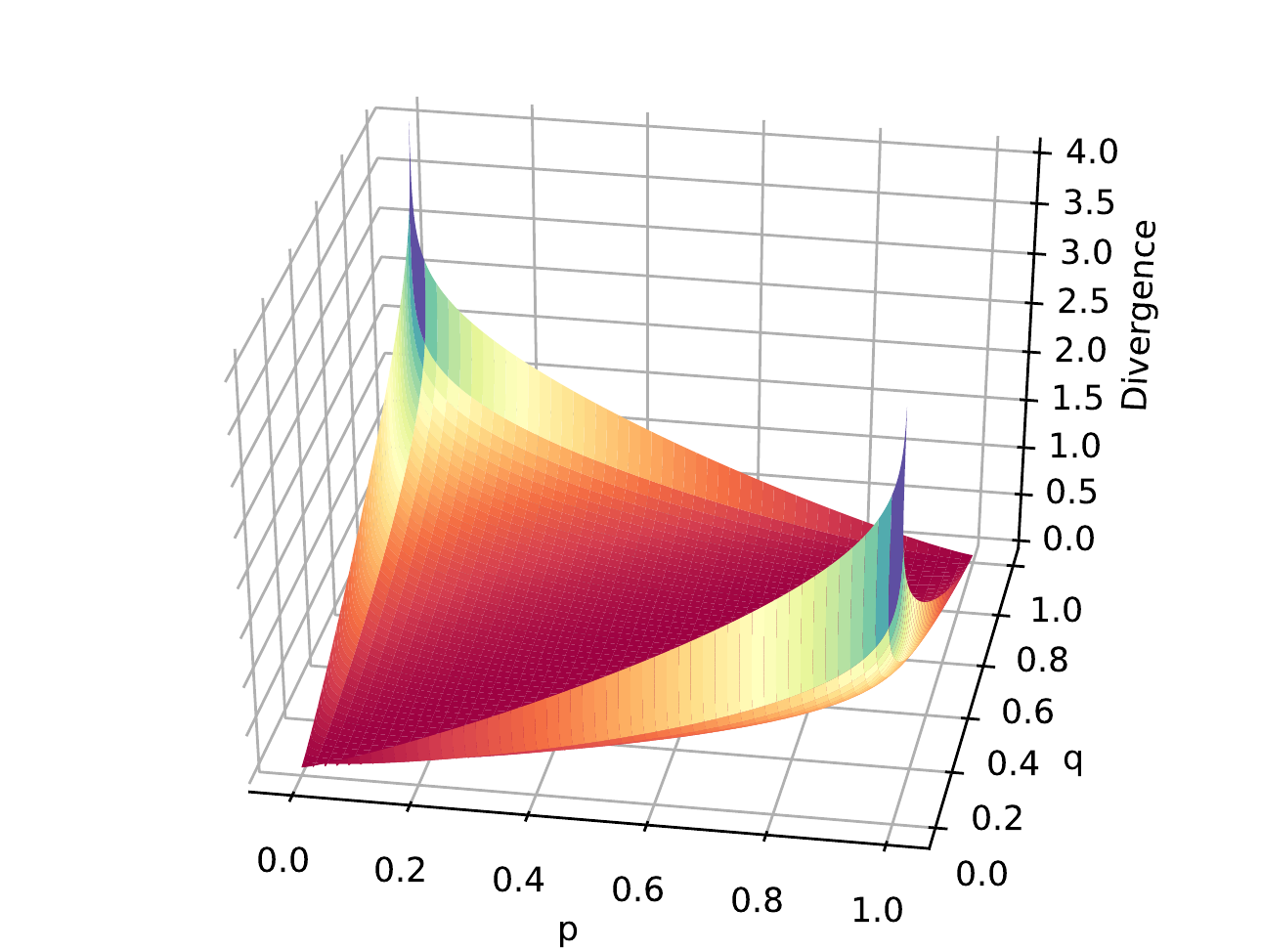} 
\caption{Symmetric KL divergence between two probability distributions with two possible outcomes $P=(p, 1-p)$ and $Q=(q, 1-q)$.}
\label{fig:kl}
\end{figure}

\begin{figure}[t]
\includegraphics[width=.95\linewidth]{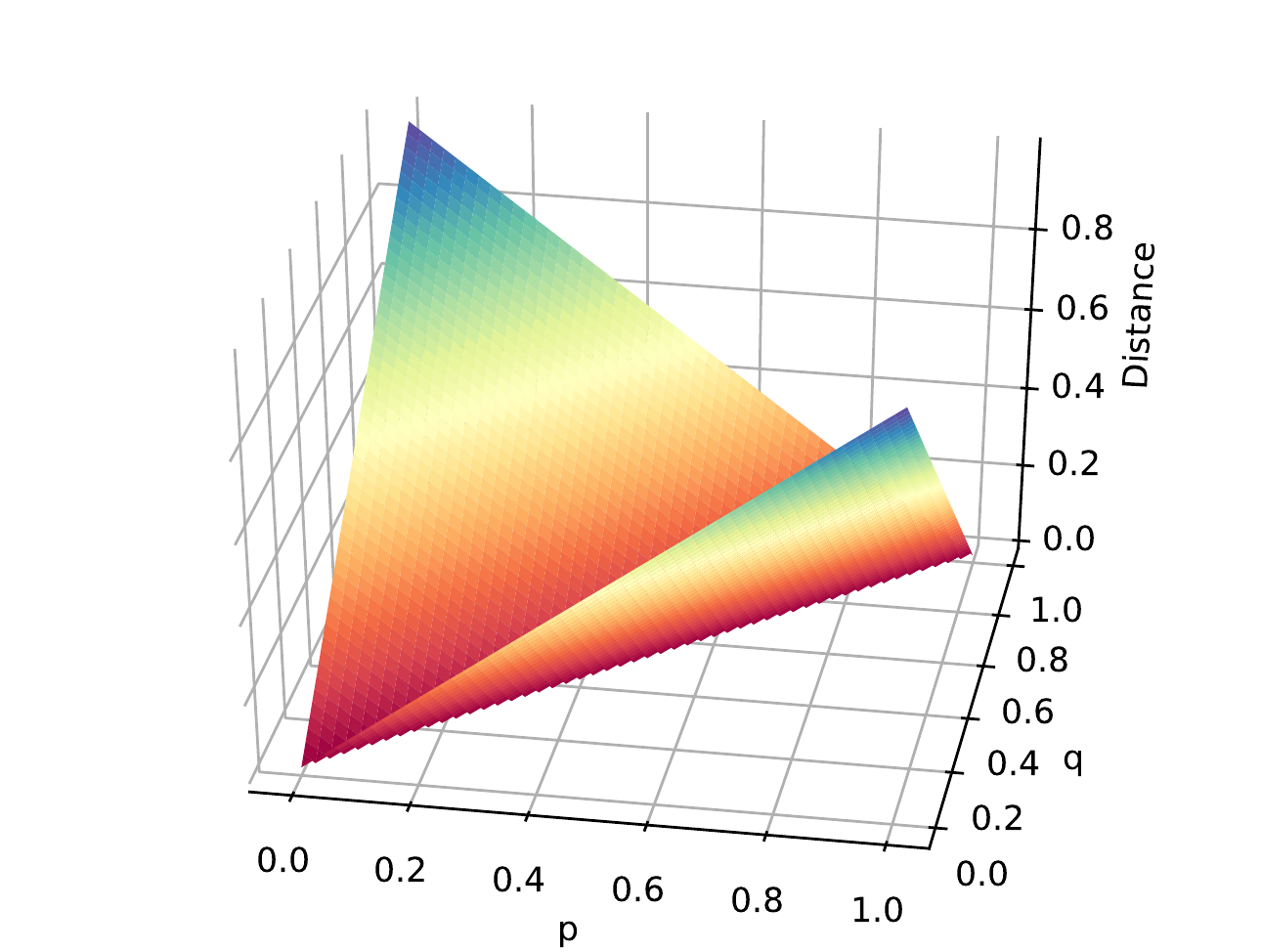} 
\caption{Wasserstein distance between two probability distributions with two possible outcomes $P=(p, 1-p)$ and $Q=(q, 1-q)$.}
\label{fig:w}
\end{figure}

\begin{table}[t]
\begin{minipage}{.5\linewidth}
\centering
\begin{tabular}{|c|c|c|}
\hline 
\textbf{s} & \textbf{P(s)} & \textbf{Pr(s)} \\ 
\hline 
s1 & 0. 97 & 0.99 \\ 
\hline 
s2 & 0.03 & 0.62 \\ 
\hline 
\end{tabular} 
\subcaption{Breakpoint 1}
\label{tab:bkpt_comparison1}
\end{minipage}%
\begin{minipage}{.5\linewidth}
\centering
\begin{tabular}{|c|c|c|}
\hline 
\textbf{s} & \textbf{P(s)} & \textbf{Pr(s)} \\
\hline 
s1 & 0.54 & 0.92 \\ 
\hline 
s2 & 0.47 & 0.94 \\ 
\hline 
\end{tabular} 
\subcaption{Breakpoint 2}
\label{tab:bkpt_comparison2}
\end{minipage}
\caption{Two candidate breakpoints, each creating two symbols (s1 and s2). The KL divergence will give a better score to breakpoint 1 while the Wasserstein distance will give a better score to breakpoint 2.}
\label{tab:bkpt_comparison}
\end{table}

\begin{figure}[t]
\includegraphics[width=\linewidth]{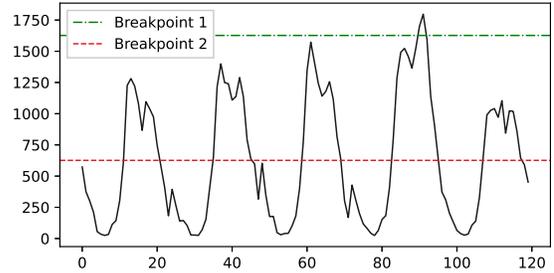} 
\caption{Two candidate breakpoints, each creating two symbols (s1 and s2).}
\label{fig:bkpt_comparison}
\end{figure}

\section{Persistence score}
The persistence score measures the persisting behavior of the symbols created by the discretization.
It is based on the Kullback-Leibler (KL) divergence.
The KL divergence \cite{kullback_information_1951} measures how a probability distribution $P$ is different from another probability distribution $Q$. 
For discrete probability distributions defined on $\mathcal{X}$, the KL divergence is defined as follows:
\begin{equation*}
D_{KL}(P || Q) = \sum_{x \in \mathcal{X}}P(x) \log(\frac{P(x)}{Q(x)}) 
\end{equation*}
This divergence is not symmetric ($D_{KL}(P || Q) \neq D_{KL}(Q || P)$). 
This is why Mörchen and Ultsch use a symmetric version obtained as follows:
\begin{equation*}
SKL(P, Q) = \frac{1}{2} (D_{KL}(P || Q) + D_{KL}(Q || P))
\end{equation*}

In Persist, the probability distributions $P$ and $Q$ are based on the probability of appearance of the symbols ($P(s)$) and their probability of repetition ($P_r(s)$): $P=(P(s), 1-P(s))$ and $Q=(P_r(s), 1-P_r(s))$.
The persistence score is computed as follows: 
\begin{equation*}
    \text{Persistence}(s) = \text{sgn}(P_r(s)-P(s)) \text{SKL}(P, Q)
\end{equation*}
The first element of the equation ($\text{sgn}(P_r(s)-P(s))$) allows to favor only the cases when the probability of repetition is superior to the probability of appearance, otherwise, it contributes negatively to the persistence score.

The symmetric KL divergence between two probability distributions with two possible outcomes as in our case is represented in Figure~\ref{fig:kl}. One of the properties of the KL divergence is that it has no upper bound, a property inherited by the persistence score. The shape of the surface produced by this divergence is also particular. The symmetric KL divergence is null when the probability distributions are equal and increases non-linearly as the difference between the distribution grows. To achieve a high value of symmetric KL, $p$ or $q$ (i. e. $P_r(s)$ or $P(s)$) have to be close to 0 or 1.
The direct consequence of these observations is that the persistence score based on the KL divergence will focus on extreme cases. 
Table~\ref{tab:bkpt_comparison} and Figure~\ref{fig:bkpt_comparison} illustrate this phenomenon.  
In this example, at the beginning of the algorithm, the first breakpoint will be selected to create two symbols.
Two candidate breakpoints are examined. 
The first breakpoint (Table \ref{tab:bkpt_comparison1}) will create a first symbol that covers almost the entire discretized time series and thus has a probability of appearance and repetition close to 1, and a second symbol that almost never appears and doesn't show a particularly recurring behavior.
The second breakpoint (Table \ref{tab:bkpt_comparison2}) will create two symbols about equally probable and with very high probabilities of repetition (greater than 0.90). 
Persist based on the KL divergence will choose breakpoint 1.
The discretized version of the time series in Figure~\ref{fig:bkpt_comparison} will consist of the succession of about 90 ``s1'', then a few ``s2'' and again ``s1'' until the end, while it would have consisted of an alternation of persistent ``s1'' and ``s2'' if breakpoint 2 had been chosen.

\section{Improving Persist}
The Wasserstein distance \cite{kantorovich_mathematical_1960}, also called Kantorovitch distance, Kantorovitch - Rubinstein distance, or earth mover's distance, is another measure of difference between probability distributions. 
It corresponds to the minimal cost to transform a distribution $P$ in another distribution $Q$ in the same space. The Wasserstein $p$-distance between two probability distributions $P$ and $Q$ is defined by the following equation where 
$\Gamma(P, Q)$ are all the possible joint distributions for $(X, Y)$ with marginal probability distributions $P$ and $Q$:
\begin{equation*}
    W_p(P,Q)=\inf_{\gamma \in \Gamma(P,Q)} (\mathbb{E}_{(x,y)\sim \gamma} d(x,y)^p)^{1/p}
\end{equation*} 

In the case of discrete probability distributions with only two possible outcomes, the Wasserstein distance becomes a simple subtraction and is defined as follows:
\begin{equation*}
W(P, Q) = | P(x_1)-Q(y_1) |
\end{equation*}

This distance is symmetric, bounded, easier to compute than the KL divergence, and it increases linearly as the difference between the distributions grows (Figure~\ref{fig:w}). Therefore, we use the Wasserstein distance to measure how the probability of appearance of the symbols and their probability of repetition are different in the score of persistence in place of the KL divergence:
\begin{equation*}
    \text{Persistence}_W(s) = \text{sgn}(P_r(s)-P(s)) \text{W}(P, Q)
\end{equation*}

In front of the choice presented in table \ref{tab:bkpt_comparison}, the persistence score computed with the KL divergence will be higher for breakpoint 1 while the persistence score computed with the Wasserstein distance will be higher for breakpoint 2. The Wasserstein distance leads here to a discretized time series with more persisting symbols, better respecting the initial intuition of the persistence score.

Finally, the initialization of the candidate breakpoints based on an equal frequency (EF) binning allows to have more possible breakpoints in high-density regions. 
However, some time series such as electrocardiograms have a structure that could be missed with this kind of binning. 
In such cases, an equal-width (EW) binning can be preferable.
It is then important to let the user choose in function of the structure of its data.

We re-implemented Persist (originally coded for MATLAB) in Python with the possibility to choose between the KL divergence and the Wasserstein distance, and between an equal-frequency or equal-width binning. It is available online\footnote{Link to the repository of Persist re-implementation in Python: https://gitlab.inria.fr/lcornang/persist\_discretization}.

\section{Experiments}
The experiment is in two parts. First, we want to evaluate the raw discretization quality provided by Persist.
Then, we look at its qualities to discretize time series for dynamical systems modeling in the form of discrete event models.

\begin{figure}[t]
\centering
\includegraphics[width=.65\linewidth]{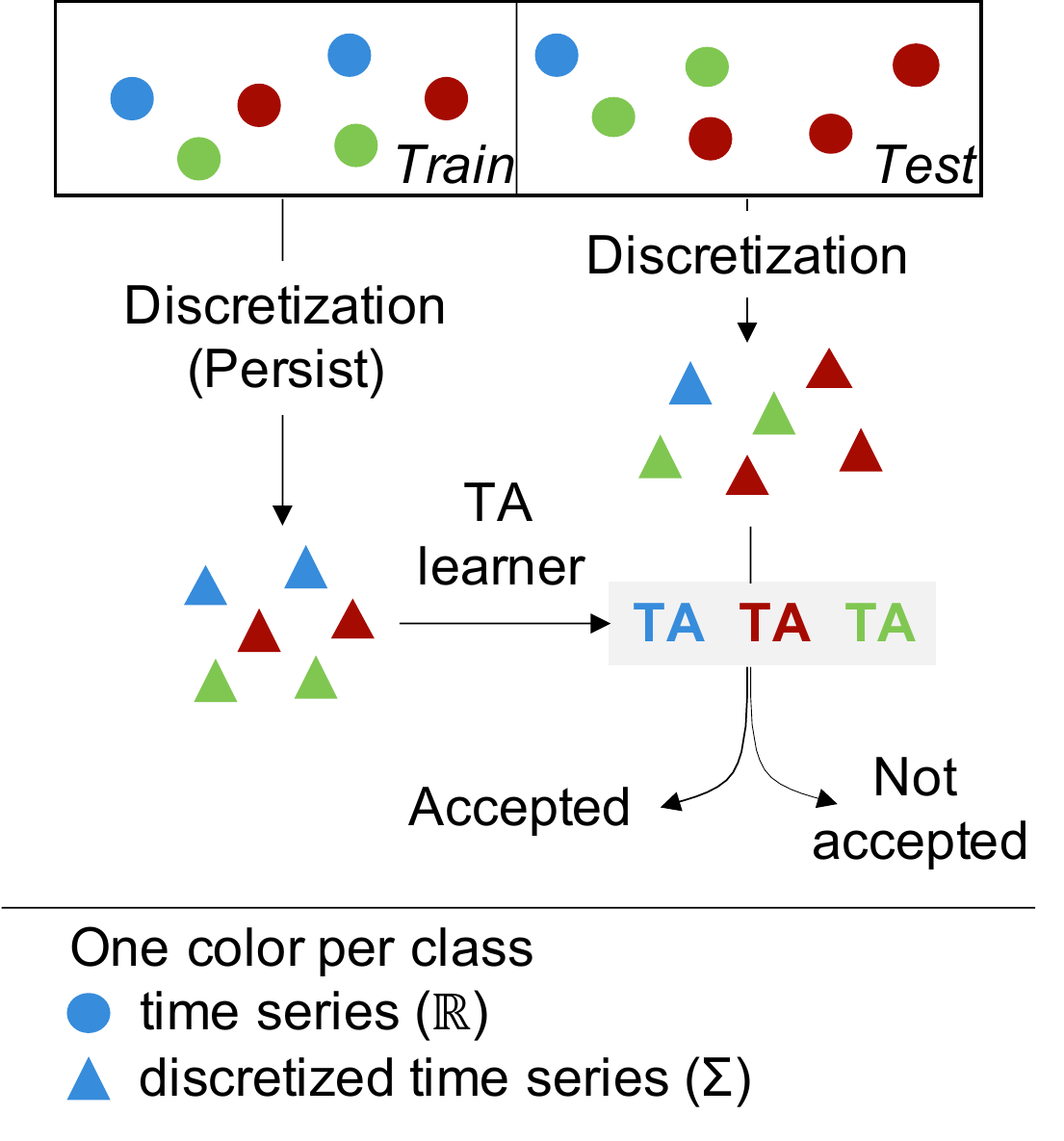} 
\caption{Time series classification using timed automata learned after a discretization step.}
\label{fig:xp}
\end{figure}

\subsection{Experimental setups}
We first want to measure the information retained in the data after the discretization. 
To allow a quantitative evaluation, we choose to evaluate the discretization through a classification task. If a good part of information is retained in the discretized time series, the classification performance should be high. 
We performed this evaluation on 111 datasets of the Time Series Classification Repository\footnote{Anthony Bagnall, Jason Lines, William Vickers and Eamonn Keogh, The UEA \& UCR Time Series Classification Repository, www.timeseriesclassification.com} (univariate datasets only).
For each dataset, Persist has produced a set of breakpoints from the train subset, used for the discretization. 
We trained a Random Forest classifier with 100 trees with the discretized train subset. 
The classification was then performed on the discretized test subset.
We measured the classification performance using the accuracy, i.e. the rate of good classification.
We tested Persist using either the KL divergence or the Wasserstein distance, and either an equal frequency binning or an equal width binning for the candidate breakpoints initialization.
We also performed the experiment using SAX to have a performance reference. 
Unlike Persist, a number of symbol must be given for SAX. We used a number of symbols ranging from 2 to 10 and a time interval width of 2 and we report all these results.

We are interested in Persist to obtain a discrete event model of the dynamical system at the origin of the time series.
Hence, to evaluate its improvement, we need to measure how good the models are.
As discrete event model, we choose timed automata \cite{alur_theory_1994} which is a common and well-studied formalism for dynamical systems. 
Timed automata (TA) are used to model systems in which time influences the behavior. A timed automaton defines states connected by transitions and the transitions from one state to another are conditioned by events and timing constraints.
The model can then be used for many purposes such as to check properties of the system (e.g. safety properties), or to perform anomaly detection in new data.
The modeling of a system by a timed automaton can be realized thanks to expert knowledge or automatically from execution data of the system. If the execution data takes the form of logs, a learning process can be applied to produce a timed automaton where the transitions labels correspond to the events present in the logs.
However, if the execution takes the form of time series (e.g. sensor data), a pre-processing step is needed to identify events in the numerical data, the discretization.
Figure~\ref{fig:xp} illustrates the experimental setup.
As in the first experiment, Persist and SAX were used to obtain the discretized time series.
However, instead of using the discretized data to train a classifier, it was here used to learn one discrete event model per class.
For each class, the corresponding discretized train time series were given to a timed automata learner called TAG \cite{cornanguer_tag_2022}, which produced a timed automaton accepting all the input sequences (i.e. there exists a path in the automaton for the sequence). 
Then the discretized test time series were injected in the timed automata. 
Each automaton received the discretized test time series of its class and as many discretized test time series of other classes. An automaton should accept the data of its own class and reject the others.
The accuracy corresponds to the good acceptance rate for the automaton.
The Time Series Classification repository gathers time series of various types (motion, sensor, traffic, image, spectrographs...). The image type, as spectrographs, differs from the others as it consists of shapes converted into pseudo time series. As it doesn't belong to the problem of modeling dynamical systems, these datasets were excluded from the experiment. 

\begin{figure}[t]
\centering
\includegraphics[width=\linewidth]{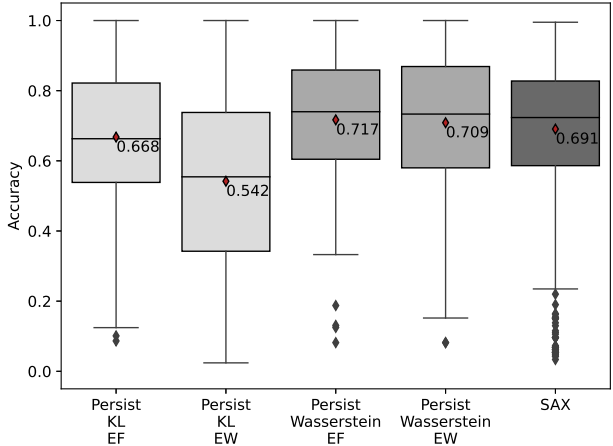} 
\caption{Classification accuracy with random forest for the different discretization strategies. The diamond indicates the mean value. EF: equal-frequency, EW: equal-width.}
\label{fig:rf_res}
\end{figure}

\begin{figure}[t]
\centering
\includegraphics[width=\linewidth]{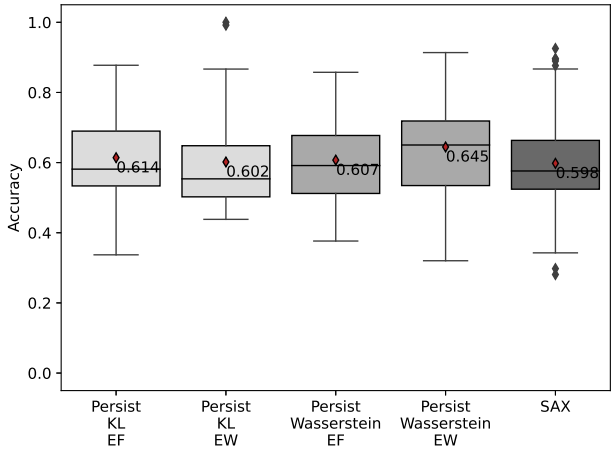} 
\caption{Classification accuracy with timed automata for the different discretization strategies.}
\label{fig:classif_at_res}
\end{figure}

\begin{figure}
    \centering
    \begin{subfigure}{\linewidth}
        \includegraphics[width=.95\linewidth]{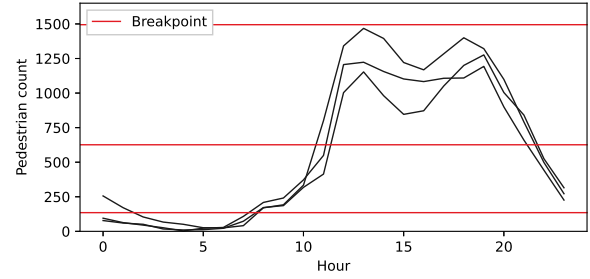} 
        \caption{Weekdays.}
        \label{fig:chinatown_bkpts_weekday}
    \end{subfigure}
    \begin{subfigure}{\linewidth}
        \includegraphics[width=.95\linewidth]{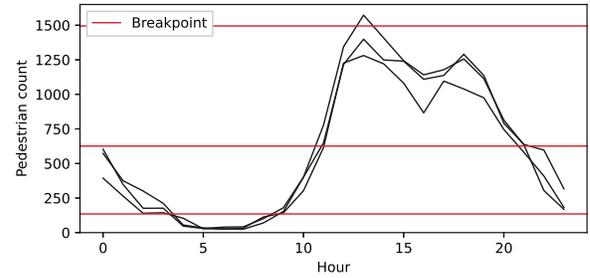} 
        \caption{Weekends.}
        \label{fig:chinatown_bkpts_weekend}
    \end{subfigure}
    \caption{Instances of time series from the Chinatown dataset and breakpoints selected by Persist (on the whole train set) with the Wasserstein distance and an equal-width binning.}
    \label{fig:chinatown_bkpts}
\end{figure}

\begin{figure}[t]
\centering
\includegraphics[width=.75\linewidth]{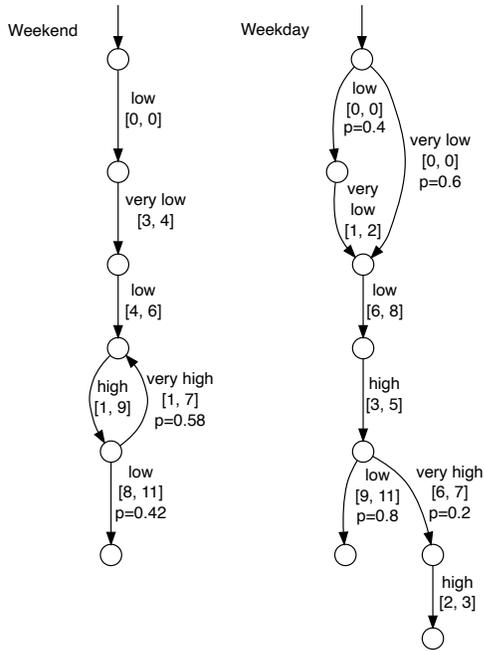} 
\caption{Discrete event model learned for each class of the Chinatown dataset.}
\label{fig:chinatown_at}
\end{figure}

\begin{table}[t]
\centering
\begin{tabular}{|l|c|}
\hline 
\textbf{Discretization method} & \textbf{Accuracy} \\\hline 
Persist (KL,  EF) & 0.686\\\hline 
Persist (Wasserstein, EF) & 0.687\\\hline 
Persist (KL, EW) & 0.780\\\hline 
Persist (Wasserstein, EW) & 0.819\\\hline 
SAX & 0.652\\\hline 
\end{tabular} 
\caption{Classification accuracy with timed automata for the Chinatown dataset.}
\label{tab:chinatown_scores}
\end{table}

\subsection{Results}
We first present the results for the classification task using Random Forest.
Figure~\ref{fig:rf_res} displays the accuracy achieved according to the discretization method.
The classification performance is increased by the replacement of the Kullback-Leibler divergence by the Wasserstein distance. 
The initialization of the candidate breakpoints by an equal frequency binning leads generally to a better performance, however, it depends on the dataset which confirms our hypothesis. 
Thanks to the Wasserstein distance, using Persist globally leads to better results than using SAX in this setup. This indicates that Persist using the Wasserstein distance allows a good information retention in the discretized data.

We now move on to the results of the second experiment.
Using automata to perform a classification task is unusual and not optimal. Indeed, each automaton is meant to represent a normal global behavior. There is no emphasis for the modeling on what makes the data of the different classes singular. For this reason, we cannot expect as good performances as while using a real classifier. 
Nevertheless, it is interesting to compare the classification performance according to the discretization method. 
If the discretization method is pertinent for discrete event modeling, a good part of the information contained in the time series would be retained in the models and therefore leading to good classification performance.
Figure~\ref{fig:classif_at_res} displays the classification accuracy using timed automata. 
When using SAX, the classification performance suffers the most from the use of timed automata.
Persist, in particular while using the Wasserstein distance and an equal-width binning, preserves a better classification accuracy.
This confirms the interest in using an improved version of Persist to preprocess time series in order to obtain a discrete event model of a dynamical system.

To provide an insight into what can be obtained with this method, we show the discretization and the discrete event models obtained for one dataset (Chinatown dataset). It consists of the pedestrian traffic along the day in a street of Melbourne. The goal is to classify the days between weekend and weekday. Figure~\ref{fig:chinatown_bkpts} shows instances of time series from this dataset.
The best accuracy using timed automata for the classification was obtained using the breakpoints from Persist with the Wasserstein distance and an equal-width binning (accuracy results in Table~\ref{tab:chinatown_scores}).
These breakpoints are shown in Figure~\ref{fig:chinatown_bkpts} and the intervals they create can be associated with a symbol (very low to very high).
Figure~\ref{fig:chinatown_at} displays the timed discrete event models obtained with the timed automata learner for each class.
A circle represents a state and the transitions from one state to another are labeled with a symbol, an interval of accepted delay since the last event, and a probability.
One can note that the activity in the street in generally higher during the night (until 3 or 4 a.m.) on weekends than on weekdays. The street also shows a more pronounced affluence during the weekend than in the weekdays afternoons. On weekdays, the end of the day is either calm, or more animated than during weekend days (with a lower probability, so probably one specific day of the week).

\section{Conclusion}
This work studies Persist, a state-of-the-art discretization algorithm originally conceived as pre-processing step for pattern mining in time series. 
Persist is based on the notion of persistence which is interesting for the modeling of dynamical systems in the form of discrete event models. 
We replaced the metric used to compute the score of persistence, originally the Kullback-Leibler divergence, with the Wasserstein distance, to avoid an undesirable over-emphasis on the extreme cases. 
We also suggested a different initialization strategy for the algorithm.
Our experiments based on a classification task have shown that the metric substitution enables a better information retention in the discretized time series, and that Persist is better suited than the state-of-the-art symbolic data representation SAX when the purpose of the discretization is to learn a model formalized as discrete event systems.
Classification is not the most common use of discrete event systems and future work will thus focus on applications for which discrete event systems are usually used such as anomaly detection or model-checking. 
Finding a criterion to determine automatically the best binning for the data would also be convenient.
Another perspective could be to associate the persistence score with other quality scores such as the reconstruction error.

\bibliography{references}

\end{document}